\documentclass[10pt,twocolumn,letterpaper]{article}
\usepackage[pagenumbers]{iccv}
\definecolor{iccvblue}{rgb}{0.21,0.49,0.74}
\usepackage[pagebackref,breaklinks,colorlinks,allcolors=iccvblue]{hyperref}
\usepackage{multirow}
\usepackage{threeparttable}
\usepackage{booktabs}  
\usepackage{subcaption}
\usepackage{algorithm}
\usepackage{algpseudocode}
\usepackage{amsmath}

\title{Improving Autoregressive Image Generation through \\ Coarse-to-Fine Token Prediction }

\author{
  Ziyao Guo$^{1}$\ \ \ \ Kaipeng Zhang$^{2,*}$\ \ \ \ Michael Qizhe Shieh$^{1,*}$\\
  {\small$^{1}$National University of Singapore \ \ \ $^{2}$Shanghai\:AI\:Lab}\\
}

\begin{document}
\maketitle
\renewcommand{\thefootnote}{*}
\footnotetext{Corresponding authors.}
\renewcommand{\thefootnote}{\arabic{footnote}}
\begin{abstract}
Autoregressive models have shown remarkable success in image generation by adapting sequential prediction techniques from language modeling. However, applying these approaches to images requires discretizing continuous pixel data through vector quantization methods like VQ-VAE.
To alleviate the quantization errors that existed in VQ-VAE, recent works tend to use larger codebooks.
However, this will accordingly expand vocabulary size, complicating the autoregressive modeling task. 
This paper aims to find a way to enjoy the benefits of large codebooks without making autoregressive modeling more difficult. 
Through empirical investigation, we discover that tokens with similar codeword representations produce similar effects on the final generated image, revealing significant redundancy in large codebooks.
Based on this insight, we propose to predict tokens from coarse to fine (CTF), realized by assigning the same coarse label for similar tokens. 
Our framework consists of two stages: (1) an autoregressive model that sequentially predicts coarse labels for each token in the sequence, and (2) an auxiliary model that simultaneously predicts fine-grained labels for all tokens conditioned on their coarse labels.
Experiments on ImageNet demonstrate our method's superior performance, achieving an average improvement of 59 points in Inception Score compared to baselines. Notably, despite adding an inference step, our approach achieves faster sampling speeds.
Code and model weights will be released in \url{https://github.com/GzyAftermath/CTF}.
\end{abstract}    
\section{Introduction}
\label{sec:intro}
Autoregressive models have demonstrated remarkable performance in language modeling by sequentially predicting the next token in a sequence \cite{radford2018improving, brown2020language, touvron2023llama}. Recently, this ``next token prediction" approach has been successfully extended to image generation tasks \citep{var, llamagen, li2024autoregressive}, achieving results that surpass diffusion-based methods on several benchmarks \citep{song2019generative, ho2020denoising, dhariwal2021diffusion, lu2022dpm, rombach2022high}.

The application of autoregressive modeling to images presents a fundamental challenge: unlike language data, which is inherently discrete, images consist of continuous pixel values. Vector quantization techniques such as VQ-VAE \citep{vqvae} address this challenge through a two-step process: first, an encoder network compresses images into lower-dimensional feature maps; second, vector quantization discretizes these features by mapping each continuous feature vector to the nearest codeword in a learned codebook. This enables representing continuous images as discrete token sequences, which are suitable for autoregressive generation.

Despite the effectiveness of VQ-VAE, its discretization process introduces quantization errors that limit reconstruction quality and, consequently, the fidelity of generated images. To alleviate this issue, recent approaches have dramatically increased codebook sizes—from 1,024 entries \citep{vqvae, vqvae2} to 16,384 \citep{llamagen}, and even up to 262,144 \citep{yu2023language, luo2024open}. While larger codebooks enhance reconstruction accuracy, they significantly complicate the autoregressive generation process since the vocabulary size is expanded accordingly, making the prediction task considerably more difficult. As a result, overall generation quality may not improve proportionally and can sometimes deteriorate \citep{yu2023language}.

\begin{figure*}[t]
    \centering
    \includegraphics[width=0.9\textwidth]{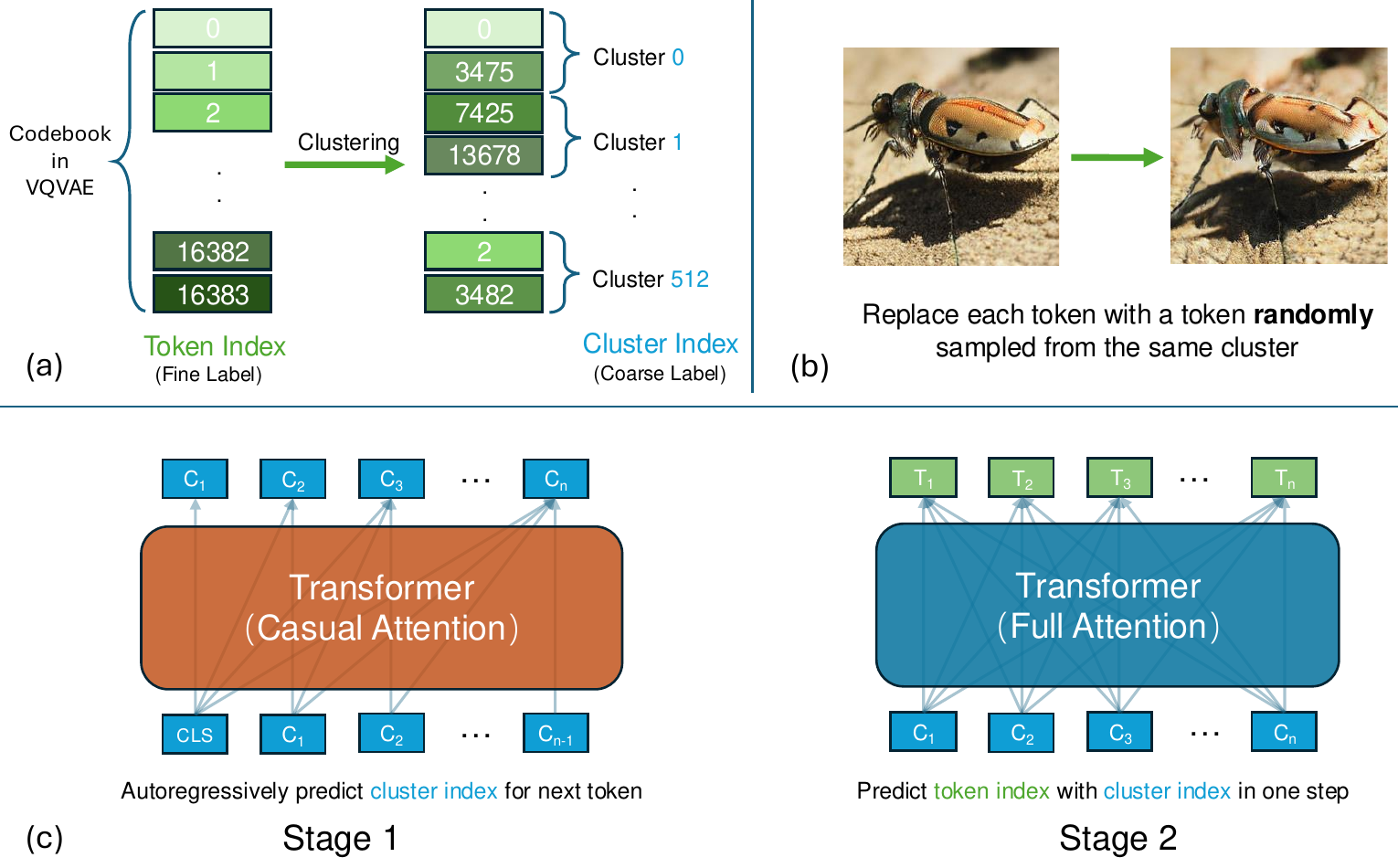}
    \caption{
    (a) The codeword clustering process, where token indices are grouped based on the similarity of their corresponding feature vectors in the codebook. 
    (b) Visual demonstration of token redundancy: replacing each token with another randomly sampled from the same cluster produces images with only minor variations in detail, preserving the overall structure and content. 
    (c) Illustration of our two-stage generation process: in the first stage, the model autoregressively predicts coarse labels (cluster indices) for each token in the sequence; then the second stage model predicts fine labels (indices in the codebook) for all tokens in a single step.
    }
    \label{fig: main}
\end{figure*}

This creates a fundamental tension: can we enjoy the benefits of large codebooks for high-quality reconstruction while keeping the complexity of autoregressive modeling tasks manageable? To address this question, we first investigate whether exact token-level predictions are as essential in image generation as they are in language modeling. Unlike language, where precise token predictions are critical for coherence, we observe that substituting image tokens with others corresponding to similar codewords results in only minor visual differences (Figure \ref{fig: main}(b)). This insight reveals significant redundancy in large codebooks—tokens with similar codeword representations produce similar effects on the generated image, yet current methods treat these closely related tokens as entirely distinct classes. This unnecessary complexity explains why even state-of-the-art autoregressive image generation methods \citep{var} with billions of parameters achieve less than 7\% accuracy in token prediction on the ImageNet validation set.

Based on our findings, we propose a novel coarse-to-fine (CTF) generation approach that leverages the visual similarity between tokens. Rather than discriminating among tens of thousands of distinct tokens, we first group tokens with similar codewords using k-means clustering and assign each cluster a corresponding coarse label. Our generation process is then divided into two stages: (1) an autoregressive model sequentially predicts the coarse label (cluster index) for each token, and (2) conditioned on these coarse labels, an auxiliary model simultaneously predicts the fine label (original codebook index) for all tokens.

This coarse-to-fine strategy effectively solves our central challenge by maintaining large codebook sizes for quality reconstruction while simplifying the autoregressive modeling task. The effective vocabulary size is reduced from the total number of tokens to a much smaller number of clusters, streamlining the learning process. Furthermore, our experiments demonstrate that once coarse labels are determined, predicting fine labels is relatively straightforward—the network can generate fine labels for all tokens in a single step, enabling our architecture to achieve both improved performance and high efficiency.

Extensive experiments confirm that our approach markedly outperforms baseline methods, achieving up to a 1-point reduction in FID scores while improving the Inception Score by an average of 59 points. Notably, despite introducing an auxiliary network and an additional inference step, our method achieves faster sampling speeds in practice due to the reduced vocabulary space.
Overall, our main contributions are:
\begin{itemize}
\item A coarse-to-fine token prediction framework that alleviates vocabulary redundancy in autoregressive image generation.
\item A systematic method for assigning coarse labels through k-means clustering of codebook vectors.
\item Empirical evidence that fine labels for all tokens in a sequence can be efficiently predicted in a single step when conditioned on coarse labels.
\item A two-stage generation algorithm that can be seamlessly integrated with various autoregressive image generation methods, yielding both improved performance and faster sampling speed.
\end{itemize}

\section{Preliminary}
\label{sec:formatting}

\subsection{VQ-VAE}
\label{sec:vqvae}
Vector Quantized Variational Autoencoder (VQ-VAE) provides a framework for mapping continuous images into a discrete latent space. Given an image $\mathbf{x} \in \mathbb{R}^{H \times W \times 3}$, the encoder $\mathcal{E}$ transforms it into a latent feature map:
\begin{equation}
\mathbf{z} = \mathcal{E}(\mathbf{x}) \in \mathbb{R}^{h \times w \times d}
\end{equation}

Each feature vector $\mathbf{z}_{ij} \in \mathbb{R}^d$ is then quantized by mapping it to its closest codeword in a learned codebook $\mathcal{B} = \{\mathbf{e}_k\}_{k=1}^K$, with the quantization defined as:
\begin{equation}
\mathbf{z}_{ij}^q = \mathbf{e}_{k^*}, \quad \text{where } k^* = \arg\min_{k \in \{1, 2, \ldots, K\}} \|\mathbf{z}_{ij} - \mathbf{e}_k\|_2^2.
\end{equation}

The quantized feature map $\mathbf{z}^q$ is then flattened into a sequence of discrete tokens $\{T_1, T_2, \ldots, T_N\}$, with $N = h \times w$, where each token $T_i \in \{1, 2, \ldots, K\}$ represents the index of the corresponding codeword in the codebook $\mathcal{B}$.

To reconstruct an image from this token sequence, tokens are first mapped to their corresponding codewords $\{\mathbf{e}_{T_1}, \mathbf{e}_{T_2}, \ldots, \mathbf{e}_{T_N}\}$, which are then reshaped to recover the quantized latent feature map $\mathbf{z}^q$. Finally, the decoder $\mathcal{D}$ reconstructs the image:
$
\hat{\mathbf{x}} = \mathcal{D}(\mathbf{z}^q).
$

\subsection{Autoregressive Image Generation}
\label{sec:autoregressive}
Autoregressive image generation leverages the discrete token representation produced by the VQ-VAE. After an image is encoded and quantized into a token sequence $\{T_1, T_2, \ldots, T_N\}$, an autoregressive model $f_\theta$ (typically based on a transformer architecture) is trained to model the conditional distribution of each token given all previous tokens in the sequence:
\begin{equation}
P_\theta(T_i \mid T_{<i}) = f_\theta(T_{<i}).
\end{equation}

The training objective is to maximize the likelihood of the observed token sequences, which is equivalent to minimizing the negative log-likelihood:
\begin{equation}
\mathcal{L}_{\text{AR}} = -\sum_{i=1}^N \log P_\theta(T_i \mid T_{<i}).
\end{equation}
During inference, the model generates the token sequence autoregressively by sampling each token from the conditional distribution $P_{\theta}(T_i \mid T_{<i})$. Once the complete token sequence is generated, it is transformed back into an image using the VQ-VAE decoder as described in Section~\ref{sec:vqvae}.

\section{Motivation and Findings}
\label{sec:motivation}
As we have introduced above, autoregressive image generation typically requires using VQ-VAE to convert images into token sequences.
To minimize the quantization error that existed in VQ-VAE, recent approaches have dramatically scaled up codebook sizes—from the original 1,024 entries \citep{vqvae, vqvae2} to 16,384 \citep{llamagen}, and even reaching 262,144 \citep{yu2023language, luo2024open}. While these larger codebooks successfully reduce quantization errors and improve reconstruction quality, they simultaneously complicate autoregressive modeling since the vocabulary size is expanded accordingly. This tension raises a crucial question: Can we maintain the benefits of larger codebooks while keeping the vocabulary complexity manageable for autoregressive models?

To answer this question, we first investigate whether redundancy exists in the expanded vocabulary space. We hypothesize that within large codebooks, multiple distinct tokens may represent similar visual information, suggesting an opportunity for vocabulary optimization without sacrificing generation quality.

To verify our hypothesis, we performed experiments based on LlamaGen \citep{llamagen}, which uses a 16,384-entry codebook. 
Firstly, we clustered the codewords into 2,048 distinct clusters using K-means based on their vector representations in the embedding space.
Then, after the autoregressive model generates all tokens, we replace them with random alternatives from the same cluster and use the VQ-VAE to convert them into images.
As can be observed in Figure~\ref{fig: main}(b), the resulting images maintained their overall structure, content, and quality, showing only minimal variations in visual details. This demonstrates that tokens with similar vector representations contribute comparable visual information to the final image.

This observation helps explain why even state-of-the-art autoregressive image generation methods struggle with token prediction accuracy. For example, VAR \citep{var}, despite employing a two-billion-parameter model, achieves less than 7\% accuracy in token prediction on ImageNet test set. The model's difficulty in discriminating between visually similar tokens severely complicates the learning process without proportionally improving generation quality.

Based on these insights, we propose to predict tokens from coarse to fine (CTF), which enables scaling codebook size while maintaining a manageable vocabulary for autoregressive modeling. The general workflow of our approach can be summarized as follows:

\begin{enumerate}
\item \textbf{Token Clustering}: We systematically group tokens with similar codeword representations using k-means clustering, assigning each cluster a coarse label. This reduces the effective vocabulary size from tens of thousands of distinct tokens to a much smaller number of semantically meaningful clusters.
\item \textbf{Coarse Label Prediction}: Use an autoregressive model to sequentially predict the coarse labels (cluster indices) for tokens in the sequence, significantly simplifying the prediction task while preserving the essential visual information needed for high-quality generation.
\item \textbf{Fine Label Prediction}: Conditioned on the predicted coarse labels, an auxiliary model is employed to predict the fine labels (original codebook indices) for all tokens in a single step.
\end{enumerate}

By applying CTF, we can enjoy the benefits brought by a large codebook while keeping the autoregressive modeling task tractable by focusing on a reduced set of meaningful visual clusters.

\section{Method}
\label{sec:method}
In this section, we first describe the process of clustering fine labels to obtain coarse labels, then introduce our two-stage prediction framework.
\subsection{Codeword Clustering}
\label{subsec:clustering}
Figure~\ref{fig: main}(a) illustrates our codeword clustering pipeline, which we formally describe below.
Given a VQ-VAE codebook $\mathcal{B} = \{\mathbf{e}_1, \mathbf{e}_2, \ldots, \mathbf{e}_K\}$ with $K$ codewords, we apply the k-means algorithm to partition these codewords into $M$ distinct clusters ($M \ll K$):
\begin{equation}
\mathcal{G} = \{G_1, G_2, \ldots, G_M\},
\end{equation}
where each cluster $G_m$ contains codewords with similar features based on their Euclidean distance in the embedding space. This clustering naturally induces a mapping function:
\begin{equation}
\phi: \{1, 2, \ldots, K\} \rightarrow \{1, 2, \ldots, M\},
\end{equation}
such that:
\begin{equation}
\phi(k) = m \quad \text{if} \quad \mathbf{e}_k \in G_m.
\end{equation}
This mapping function $\phi$ assigns each fine-grained token to its respective cluster, effectively creating a coarse-grained representation where visually similar tokens share the same cluster label. This process substantially reduces the vocabulary size from $K$ to $M$, making the autoregressive modeling task more tractable. In our implementation, we use a default configuration that maps 16,384 fine-grained token categories to 512 coarse-grained clusters, yielding a 32-fold reduction in vocabulary size.

\subsection{Coarse-to-Fine Prediction Framework}
\label{subsec:c2f-prediction}
As illustrated in Figure~\ref{fig: main}(c), our generation process consists of two sequential stages, predicting coarse labels and fine labels for tokens in the sequence respectively.

\subsubsection{Stage 1: Autoregressive Coarse Label Prediction}
\label{subsubsec:stage1}
In the first stage, we train an autoregressive model to predict coarse labels for tokens in a sequence to capture the overall structure and content of the image.
For each token position $i$, we define the coarse label as $C_i = \phi(T_i)$, where $T_i$ is the original fine-grained token. The autoregressive model learns to predict:
\begin{equation}
P_\theta^c(C_i | C_{<i}) = f_\theta^c(C_{<i}).
\end{equation}
The training objective is to maximize the likelihood of the correct coarse labels in the sequence:
\begin{equation}
\mathcal{L}_{\text{coarse}} = -\sum_{i=1}^N \log P_\theta^c(C_i | C_{<i}).
\end{equation}

This formulation significantly simplifies the autoregressive modeling task, allowing the model to focus on capturing meaningful structural relationships rather than discriminating among highly similar tokens.

\subsubsection{Stage 2: Parallel Fine Label Prediction}
\label{subsubsec:stage2}
After stage 1, we obtain the complete sequence of coarse labels $\mathbf{C} = \{C_1, C_2, \ldots, C_N\}$ for all tokens.
Next, we deploy a full-attention transformer model to simultaneously predict the fine-grained labels for the entire sequence.
This model takes the complete coarse sequence as input and outputs probability distributions for each fine-grained token  $\{T_1, T_2, \ldots, T_N\}$.
The training objective for this stage is to maximize the likelihood of the actual tokens conditioned on the coarse sequence:
    \begin{equation}
    \mathcal{L}_{\text{fine}} = -\sum_{i=1}^N \log p_\theta(T_i | \mathbf{C}).
    \end{equation}
    
Importantly, our experiments demonstrate that this fine-label prediction task is substantially simpler compared with the autoregressive token prediction problem. Since each coarse label $C_i$ restricts the possible fine labels to a much smaller subset (those within the corresponding cluster), the model only needs to discriminate among tokens within the same cluster rather than across the entire vocabulary. Furthermore, by predicting all fine labels simultaneously with a full-attention mechanism, the model can leverage global context from the entire coarse sequence, enabling more coherent refinement.

\subsection{Details in Training and Inference }
\label{subsec:training-inference}
A key advantage of our approach is that the Stage 1 and Stage 2 models are independent in training, thus they can be trained in parallel to improve training efficiency.
Also, our framework is agnostic to the specific architecture of the Stage 1 model—any autoregressive model can be employed, as long as its outputs are token sequences rather than continuous representations \citep{li2024autoregressive}.
This flexibility makes our method compatible with most autoregressive architectures.

For inference, we first use the Stage 1 model to autoregressively generate the coarse label sequence. 
In this process, techniques such as temperature control and classifier-free guidance~\citep{ho2022classifier} can be applied, just as in conventional autoregressive models. 
Once the coarse sequence is complete, we pass it to the Stage 2 model, which produces all fine labels in a single step.
In this process, we find that performing top-$k$ sampling also helps to improve performance, largely by increasing output diversity (see Section~\ref{sec:topk}).
The resulting fine-grained token sequence is then processed by the VQ-VAE decoder to generate the final image.

This two-stage framework effectively addresses the vocabulary redundancy problem in autoregressive image generation. By first capturing the essential structure using a reduced set of semantically meaningful labels and then recovering detailed visual information within each cluster, our approach simplifies the most challenging aspect of autoregressive prediction while preserving high reconstruction fidelity.

\section{Experiments}
\begin{table}[]
\resizebox{\columnwidth}{!}{
\begin{tabular}{c|lc|cccc}
\toprule
Type                    & Model                        & \#Para.                   & FID↓                         & IS↑                            & Pr.↑                         & Re.↑                         \\ \midrule
                        & BigGAN \citep{brock2018large}                      & 112M                      & 6.95                         & 224.5                          & 0.89                         & 0.38                         \\
                        & GigaGAN \citep{kang2023scaling}                     & 569M                      & 3.45                         & 225.5                          & 0.84                         & 0.61                         \\
\multirow{-3}{*}{GAN}   & StyleGAN-XL \citep{sauer2022stylegan}                 & 166M                      & 2.30                          & 265.1                          & 0.78                         & 0.53                         \\ \midrule
                        & ADM \citep{dhariwal2021diffusion}                         & 554M                      & 10.94                        & 101.0                            & 0.69                         & 0.63                         \\
                        & CDM \citep{ho2022cascaded}                         & -                         & 4.88                         & 158.7                          & -                            & -                            \\
                        & LDM-4 \citep{rombach2022high}                       & 400M                      & 3.60                          & 247.7                          & -                            & -                            \\
\multirow{-4}{*}{Diff.} & DiT-XL/2 \citep{peebles2023scalable}                    & 675M                      & 2.27                         & 278.2                          & 0.83                         & 0.57                         \\ \midrule
                        & MaskGIT \citep{chang2022maskgit}                     & 227M                      & 6.18                         & 182.1                          & 0.80                          & 0.51                         \\
\multirow{-2}{*}{Mask.} & MaskGIT-re \citep{chang2022maskgit}                  & 227M                      & 4.02                         & 355.6                          & -                            & -                            \\ \midrule
                        & VAR-d16 \citep{var}                     & 310M                      & 3.30                          & 274.4                          & 0.84                         & 0.51                         \\
                        & VAR-d20 \citep{var}                     & 600M                      & 2.57                         & 302.6                          & 0.83                         & 0.56                         \\
\multirow{-3}{*}{VAR}   & VAR-d24 \citep{var}           & 1.0B                      & 2.09                         & 312.9                          & 0.82                         & 0.59                         \\ \midrule
                        & VQGAN \citep{esser2021taming}                      & 227M                      & 18.65                        & 80.4                           & 0.78                         & 0.26                         \\
                        & VQGAN \citep{esser2021taming}                       & 1.4B                      & 15.78                        & 74.3                           & -                            & -                            \\
                        & VQGAN-re \citep{esser2021taming}                    & 1.4B                      & 5.20                          & 280.3                          & -                            & -                            \\
                        & ViT-VQGAN \citep{yu2021vector}                   & 1.7B                      & 4.17                         & 175.1                          & -                            & -                            \\
                        & ViT-VQGAN-re \citep{yu2021vector}                & 1.7B                      & 3.48                         & 175.1                          & -                            & -                            \\
                        & RQTran. \citep{lee2022autoregressive}                     & 3.8B                      & 7.55                         & 134.0                            & -                            & -                            \\

\multirow{-7}{*}{AR}    & RQTran.-re \citep{lee2022autoregressive}                  & 3.8B                      & 3.80                          & 323.7                          & -                            & -                            \\ \midrule
& IAR-B$^\dagger$ \citep{hu2025improving}        & 111M   & 5.14   & 202.00       & 0.85                         & 0.45                         \\
& IAR-L$^\dagger$ \citep{hu2025improving}        & 343M   & 3.18   & 234.80       & 0.82                        & 0.53                          \\
                        & IAR-XL$^\dagger$ \citep{hu2025improving}                     & 775M                      & 2.52                    & 248.10                  & 0.82                         & 0.58                           \\
                        & LlamaGen-B$^\dagger$ \citep{llamagen}        & 111M   & 6.09   & 182.54       & 0.85                         & 0.42                         \\
                        & LlamaGen-L$^\dagger$ \citep{llamagen}                     & 343M                      & 3.07                         & 256.06                          & 0.83                         & 0.52                         \\
\multirow{-6}{*}{AR}   & LlamaGen-XL$^\dagger$ \citep{llamagen}           & 775M                      & 2.62                         & 244.08                          & 0.80                         &  0.57                         \\ \midrule
                        &  \cellcolor[HTML]{c9eef6}LlamaGen-B \citep{llamagen}                  &  \cellcolor[HTML]{c9eef6}111M                      &  \cellcolor[HTML]{c9eef6}5.46                         &  \cellcolor[HTML]{c9eef6}193.61                         &  \cellcolor[HTML]{c9eef6}0.83                         &  \cellcolor[HTML]{c9eef6}0.45                         \\  
                        &  \cellcolor[HTML]{c9eef6}+ CTF &  \cellcolor[HTML]{c9eef6}87M$^{*}$ &  \cellcolor[HTML]{c9eef6}4.15 &  \cellcolor[HTML]{c9eef6}254.99  &  \cellcolor[HTML]{c9eef6}0.86     &  \cellcolor[HTML]{c9eef6}0.48     \\
                        & \cellcolor[HTML]{64ddf5}LlamaGen-L \citep{llamagen}                  & \cellcolor[HTML]{64ddf5}343M                      & \cellcolor[HTML]{64ddf5}3.80                          & \cellcolor[HTML]{64ddf5}248.28                         & \cellcolor[HTML]{64ddf5}0.83                         & \cellcolor[HTML]{64ddf5}0.52                         \\
                        & \cellcolor[HTML]{64ddf5}+ CTF & \cellcolor[HTML]{64ddf5}310M$^{*}$ & \cellcolor[HTML]{64ddf5}2.97 & \cellcolor[HTML]{64ddf5}291.53 & \cellcolor[HTML]{64ddf5}0.84 & \cellcolor[HTML]{64ddf5}0.53 \\
                        &\cellcolor[HTML]{c9eef6}LlamaGen-XL \citep{llamagen}                 &\cellcolor[HTML]{c9eef6}775M                      &\cellcolor[HTML]{c9eef6}3.39                          &\cellcolor[HTML]{c9eef6}227.08                         &\cellcolor[HTML]{c9eef6}0.81                         &\cellcolor[HTML]{c9eef6}0.54                         \\
\multirow{-6}{*}{AR}    &\cellcolor[HTML]{c9eef6}+ CTF &\cellcolor[HTML]{c9eef6}734M$^*$  &\cellcolor[HTML]{c9eef6}2.76    &\cellcolor[HTML]{c9eef6}299.69      &\cellcolor[HTML]{c9eef6}0.84    &\cellcolor[HTML]{c9eef6}0.55    \\ \bottomrule

\end{tabular}

}
\caption{Performance comparison on class-conditional ImageNet at \textbf{256×256} resolution. Models are evaluated using FID, Inception Score (IS), precision (Pr.), and recall (Re.) metrics. 
The background is colorized for convenient comparison with baseline.
Models with \textbf{$\dagger$} were trained at \textbf{384×384} resolution and downsampled to 256×256 for evaluation. \textbf{$*$}: Our autoregressive models have fewer parameters due to the reduced vocabulary size, complemented by an auxiliary network for fine-grained prediction, see Section \ref{sec: Sampling Speed} for a detailed efficiency analysis.}
\label{tab: main compare}
\end{table}
\subsection{Setup}
Our work tackles the redundancy in token labels for autoregressive image generation, thereby enabling the use of any AR method that produces token sequences~\citep{var, chang2022maskgit, llamagen}.
We adopt LlamaGen~\citep{llamagen} as our backbone, owing to its simplicity—training both the VQ-VAE and autoregressive model solely on ImageNet without using complex training or inference strategy.

\textbf{Model Architectures.} We employ the VQ-VAE trained by LlamaGen~\citep{llamagen} on ImageNet-1K as our image tokenizer.
The tokenizer has a codebook size of 16,384 and downsamples the input image by a factor of 16×16.
For both stage 1 and stage 2, we adopt the same architecture as LlamaGen~\citep{llamagen}, with only one modification: the fine stage model uses full attention instead of causal attention.

\textbf{Benchmark.}
We compare our method with previous work on the class-conditional image generation task using the ImageNet-1K benchmark~\citep{deng2009imagenet}. For fairness, all results of previous work are obtained from~\citep{llamagen, var}.

\textbf{Training Settings.}
Following LlamaGen~\citep{llamagen}, we train all models for 300 epochs using a batch size of 256. We employ the AdamW optimizer with parameters $\beta_1=0.9$, $\beta_2=0.95$, and a weight decay of 0.05 while applying gradient clipping at 1.0 to stabilize training. To enable classifier-free guidance, we apply a dropout rate of 0.1 on the class token embeddings. The learning rate is initialized at $1\times10^{-4}$ and decays to $1\times10^{-5}$ following a cosine annealing schedule.

\textbf{Sampling Settings.}
For simplicity, we adopt a consistent set of sampling hyperparameters across all model configurations. Specifically, we set the classifier-free guidance scale to 2.0, temperature to 1.1, and use unrestricted sampling with top-$k$=0.

\subsection{Main Results}
We evaluate our approach against baseline methods on the class-conditional ImageNet 256×256 benchmark, as shown in Table~\ref{tab: main compare}. Our method consistently achieves substantial performance gains across model sizes ranging from 100M to 775M parameters.

Compared to LlamaGen-B, our approach reduces the FID score by over 1.0 while significantly improving the Inception Score. Additionally, LlamaGen-L with our method surpasses LlamaGen-XL across all metrics by a large margin, highlighting the effectiveness of our coarse-to-fine prediction strategy.

Notably, despite being trained at 256×256 resolution, our models outperform baseline models trained at a higher 384×384 resolution in most cases. Overall, our approach substantially enhances image generation quality—improving the Inception Score by an average of 59 points—while also offering faster sampling speed (see Section~\ref{sec: Sampling Speed}).

\subsection{Sampling Speed}
\label{sec: Sampling Speed}
\begin{table}[]
\resizebox{\columnwidth}{!}{%
\begin{tabular}{l|ccccc}
\toprule
Model           & Total Param     & Step  & images/sec ↑ & FID ↓ & IS ↑ \\ \midrule
LlamaGen-B      & 111M      & 256   & \textbf{13.75}        & 5.46   & 193.61    \\
+ CTF  & 87M+343M  & 256+1 & 12.83       & \textbf{4.15}    & \textbf{254.99}   \\ \midrule
LlamaGen-L      & 343M      & 256   & 7.50        & 3.80    & 248.28    \\
+ CTF  & 310M+343M & 256+1 & \textbf{8.71}       & \textbf{2.97}    &  \textbf{291.53}  \\ \midrule
LlamaGen-XL     & 775M      & 256   & 5.32        & 3.39    & 227.08    \\
+ CTF & 734M+343M & 256+1 & \textbf{6.26}        & \textbf{2.76}    & \textbf{299.69}   \\ \bottomrule
\end{tabular}%
}
\caption{Sampling speed comparison, measured on a single A100 GPU with a batch size of 64. Although our method employs an auxiliary network with 343M parameters, our method achieved better sampling efficiency in most cases.}
\label{tab: throughput}
\end{table}
In Table \ref{tab: throughput}, we compare the sampling speeds of our proposed method with the baseline approach. Despite introducing an auxiliary network for detailed label prediction—which requires an additional inference step—our method achieves faster sampling speeds in most test cases.

This performance improvement stems primarily from the efficiency of our autoregressive model architecture. Unlike the baseline, our model's fully connected layer contains fewer parameters, as it only needs to predict coarse labels with significantly fewer categories than detailed labels. This advantage becomes particularly significant considering that the autoregressive model executes 256 times during the sampling process, effectively offsetting the computational cost of the auxiliary network.

Furthermore, our approach's speed advantage becomes increasingly pronounced as the autoregressive model scales up in size. The reduction in parameter count in the fully connected layer creates compounding efficiency benefits that outweigh the overhead of the additional network component.

\subsection{Training Efficiency}

\begin{figure}[t]
    \centering
    \includegraphics[width=\columnwidth]{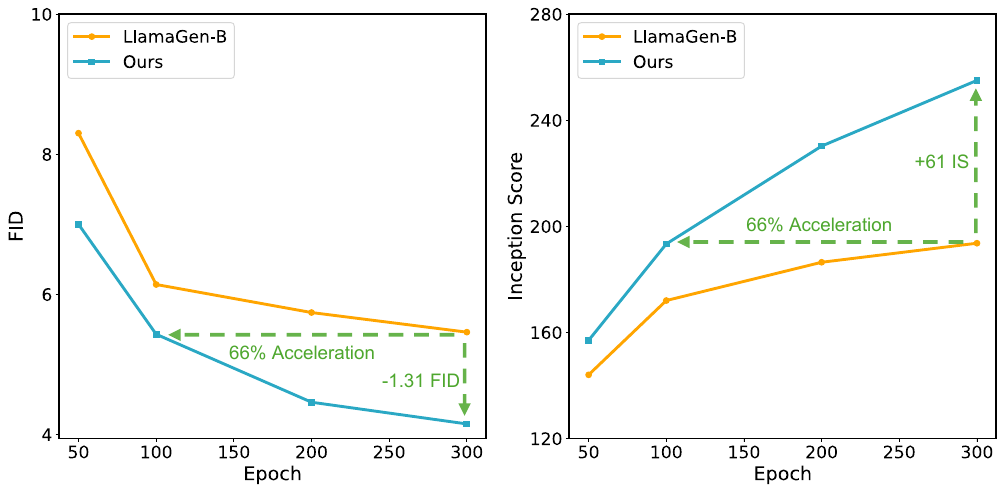}
    \caption{Model performance comparison on different epochs. When our method is applied, models achieve significantly better performance.}
    \label{fig: efficiency}
\end{figure}

To evaluate whether our coarse-to-fine prediction strategy enables more efficient learning of image token distributions, we compared model performance across different training durations. Figure~\ref{fig: efficiency} presents the FID and IS metrics for both our approach and the baseline method at various training epochs.

The results demonstrate that our method consistently outperforms the baseline by a substantial margin throughout the training process. Most notably, our model achieves an FID of 5.43 and an IS of 193.34 after just 100 epochs, comparable to the baseline model's performance after 300 epochs (FID: 5.46, IS: 193.61). This represents a 3× reduction in required training time to reach equivalent generation quality.

These findings confirm that our coarse-to-fine approach not only accelerates convergence but also enhances the model's capacity to learn meaningful token distributions. By reducing vocabulary redundancy and focusing on hierarchical relationships between tokens, our method improves both training efficiency and the quality of generated images.

\begin{table*}[htbp]
    \centering
    \resizebox{0.95\textwidth}{!}{%
    \renewcommand{\tabcolsep}{4pt} 
        \begin{minipage}{\textwidth}
            \begin{subtable}[t]{0.32\textwidth}
                \centering
                \begin{tabular}{c|cccc}
                    \toprule
                     Num. & FID$\downarrow$ & IS$\uparrow$ & Pr.$\uparrow$ & Re.$\uparrow$ \\ \midrule
                    128 & 11.23 & 98.41 & 0.68 & 0.39 \\
                    \rowcolor{gray!30}
                    512 & 5.51 & \textbf{273.45} & 0.88 & 0.44 \\
                    1024 & \textbf{5.38} & 269.10 & \textbf{0.89} & 0.44 \\
                    \bottomrule
                \end{tabular}
                \caption{Number of Clusters}
                \label{Tab: ablation on Number of clusters}
            \end{subtable}
            \hfill
            \begin{subtable}[t]{0.32\textwidth}
                \centering
                \begin{tabular}{c|cccc}
                    \toprule
                     Para. & FID$\downarrow$ & IS$\uparrow$ & Pr.$\uparrow$ & Re.$\uparrow$ \\ \midrule
                    111M & 5.51 & 273.45 & 0.88 & 0.44 \\
                    \rowcolor{gray!30}
                    343M & 5.33 & 285.15 & 0.89 & 0.43 \\
                    775M & \textbf{5.27} & \textbf{289.10} & 0.89 & \textbf{0.46} \\
                    \bottomrule
                \end{tabular}
                \caption{Model Size (stage 2)}
                \label{Tab: ablation on Model Size}
            \end{subtable}
            \hfill
            \begin{subtable}[t]{0.32\textwidth}
                \centering
                \begin{tabular}{c|cccc}
                    \toprule
                     CFG & FID$\downarrow$ & IS$\uparrow$ & Pr.$\uparrow$ & Re.$\uparrow$ \\ \midrule
                    1.75 & \textbf{4.24} & 247.02 & 0.86 & \textbf{0.48} \\
                    \rowcolor{gray!30}
                    2.00 & 5.33 & 285.15 & 0.89 & 0.43 \\
                    2.25 & 6.71 & \textbf{313.06} & \textbf{0.90} & 0.41 \\
                    \bottomrule
                \end{tabular}
                \caption{CFG}
                \label{Tab: ablation on CFG}
            \end{subtable}
            
            \vspace{1em}
            
            \begin{subtable}[t]{0.32\textwidth}
                \centering
                \begin{tabular}{c|cccc}
                    \toprule
                    Top-$k$ & FID$\downarrow$ & IS$\uparrow$ & Pr.$\uparrow$ & Re.$\uparrow$ \\ \midrule
                    1 & 6.95 & 210.21 & 0.87 & 0.39 \\
                    32 & 5.35 & \textbf{286.73} & 0.89 & 0.43 \\
                    \rowcolor{gray!30}
                    All & \textbf{5.33} & 285.15 & 0.89 & 0.43 \\
                    \bottomrule
                \end{tabular}
                \caption{Top-$k$ (stage 2)}
                \label{Tab: ablation on Topk}
            \end{subtable}
            \hfill
            \begin{subtable}[t]{0.32\textwidth}
                \centering
                \begin{tabular}{c|cccc}
                    \toprule
                     $\tau_1$& FID$\downarrow$ & IS$\uparrow$ & Pr.$\uparrow$ & Re.$\uparrow$ \\ \midrule
                    0.9 & 7.30 & \textbf{292.73} & \textbf{0.90} & 0.39 \\
                    1.0 & 5.33 & 285.15 & 0.89 & 0.43 \\
                    \rowcolor{gray!30}
                    1.1 & \textbf{4.16} & 259.45 & 0.86 & \textbf{0.47} \\
                    \bottomrule
                \end{tabular}
                \caption{Temperature (stage 1)}
                \label{Tab: ablation on Temperature (stage 1)}
            \end{subtable}
            \hfill
            \begin{subtable}[t]{0.32\textwidth}
                \centering
                \begin{tabular}{c|cccc}
                    \toprule
                    $\tau_2$ & FID$\downarrow$ & IS$\uparrow$ & Pr.$\uparrow$ & Re.$\uparrow$ \\ \midrule
                    0.9 & 5.55 & \textbf{288.01} & \textbf{0.89} & 0.43 \\
                    1.0 & 4.16 & 259.45 & 0.86 & 0.47 \\
                    \rowcolor{gray!30}
                    1.1 & \textbf{4.15} & 254.99 & 0.86 & \textbf{0.48} \\
                    \bottomrule
                \end{tabular}
                \caption{Temperature (stage 2)}
                \label{Tab: ablation on Temperature (stage 2)}
                \label{test}
            \end{subtable}
            
        \end{minipage}
    }
    \caption{Ablation Studies. Hyper-parameters with a grey background are selected as the default settings.}
    \label{tab: ablation}
\end{table*}
\subsection{Ablation}
In this section, we conduct ablation studies to investigate the impact of various hyperparameters involved in training and inference processes. The results are reported in Table~\ref{tab: ablation}.

\subsubsection{Number of Clusters.} According to the results reported in Table~\ref{Tab: ablation on Number of clusters}, performance degrades significantly when the number of clusters is low. This occurs because each cluster contains too many fine labels when clustering is sparse. For instance, with 128 clusters, each cluster contains approximately 128 fine labels on average, compared to just 32 labels per cluster when using 512 clusters. When a single cluster encompasses numerous fine labels, the auxiliary model struggles to accurately predict fine labels based on the given coarse labels in a single step.
However, simply increasing the cluster number does not always yield improvements, as the task approaches a standard autoregressive image generation process when the cluster number approaches the total number of detailed labels. Our experiments reveal a clear trade-off in this parameter. Based on experimental results, we established 512 as the optimal number of clusters for our default configuration.

\subsubsection{K-means.}
To evaluate the importance of clustering based on codeword similarity in our method, we conducted an experiment replacing K-means with random clustering. The results demonstrate that the auxiliary model fails to accurately predict fine labels from coarse labels when clustering is performed randomly. This failure can be attributed to the weak correlation between tokens within the same randomly assigned cluster, which significantly impairs the model's ability to capture inter-token relationships.
These findings confirm the effectiveness of our algorithm design, highlighting that meaningful clustering based on codeword similarity is essential for the hierarchical prediction mechanism to function properly. The structured grouping provided by K-means creates coherent clusters that enable the auxiliary model to learn meaningful patterns between coarse and fine labels.

\subsubsection{Auxiliary Model Size.}
As shown in Table~\ref{Tab: ablation on Model Size}, increasing the auxiliary model size leads to consistent performance improvements. However, we observe that this benefit becomes marginal once the parameter count approaches 343M. Balancing performance gains against computational efficiency, we selected the 343M parameter auxiliary model as our default configuration.

\subsubsection{Classifier Free Guidance.} As observed in Table \ref{Tab: ablation on CFG}, our experiments reveal an interesting trade-off: lower CFG (Classifier-Free Guidance) factors improve FID scores, while higher CFG factors enhance Inception Scores. 
For the consideration of balanced performance, we selected a CFG factor of 2 as our default setting, which performed well across various settings.

\subsubsection{Temperature.}
According to the results reported in Table~\ref{Tab: ablation on Temperature (stage 1)} and Table~\ref{tab: main compare}, our models achieve significantly better IS scores than the baseline method, demonstrating the effectiveness of our coarse-to-fine prediction strategy. However, our method's FID is relatively inferior when temperature control is not applied.

We attribute this performance gap to two key factors: (1) during the autoregressive generation process, our models can sample from only 512 classes while the baseline method can sample from 16,384 classes, and (2) our regressive model exhibits higher confidence in its predictions, as our coarse-to-fine strategy simplifies the task (our model's training loss is approximately 5 compared to the baseline's loss of around 7).

Although these factors help to improve the quality of the generated images, they reduce the diversity of the generation. However, this can be effectively addressed by using temperature control to make the model's predictions less confident.

As demonstrated in Table~\ref{Tab: ablation on Temperature (stage 1)} and Table~\ref{Tab: ablation on Temperature (stage 2)}, using a larger temperature effectively improves the diversity of the generation (reflected in the better FID score). The temperature control plays a crucial role in balancing the trade-off between generation quality and diversity in our approach.

\subsubsection{Top-k.}
\label{sec:topk}
As demonstrated in Table~\ref{Tab: ablation on Topk}, restricting the sampling to only the highest probability class (Top-$k$=1) in stage 2 leads to significantly worse performance on both FID and Inception Score metrics. This degradation occurs because these evaluation metrics assess not only image quality but also diversity. When Top-$k$ is set to 1, the generation process can only select the token with the highest probability, which substantially limits output variability. To enhance diversity in our generated images, we follow \citet{llamagen}'s recommendation and adopt Top-$k$=0 (allowing sampling across the entire probability distribution) as our default configuration.

\subsection{Visualization}
\begin{figure}[t]
    \centering
    \includegraphics[width=\columnwidth]{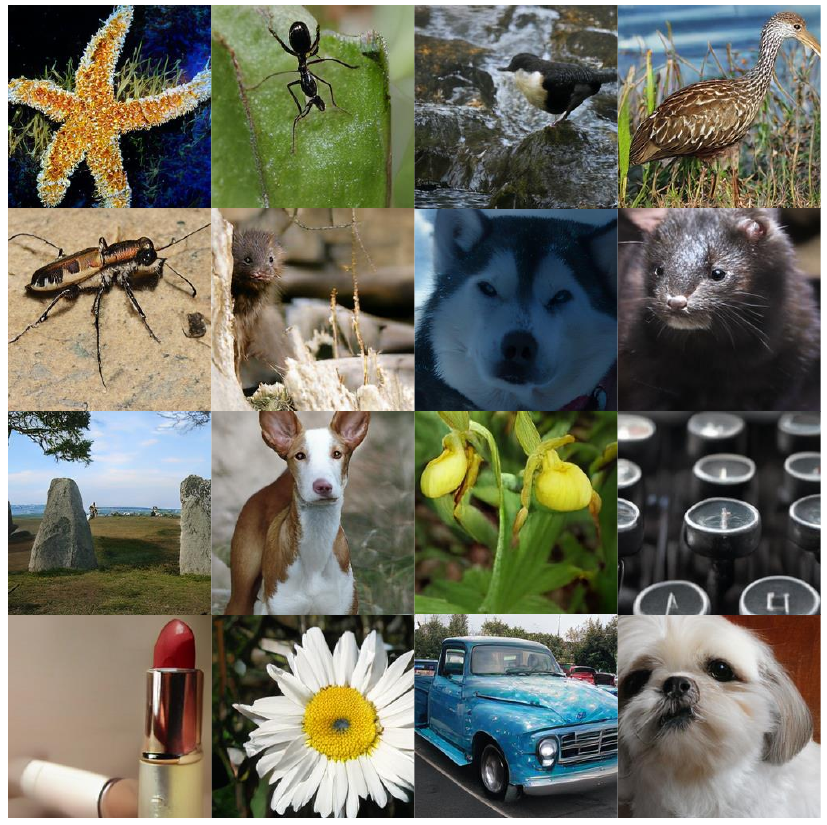}
    \caption{Generation results of our method (based on LlamaGen-XL) on ImageNet 256×256 benchmark.}
    \label{fig: visual}
\end{figure}

We visualize the images generated by our method (based on LlamaGen-XL) in Figure~\ref{fig: visual}. As shown, our model's coarse-to-fine prediction strategy produces images with remarkably detailed textures and structures. The progressive refinement successfully preserves global composition while enhancing local features, demonstrating the effectiveness of our hierarchical generation approach.
\section{Related Work}

\subsection{Large Language Models.}

Large language models (LLMs)~\citep{devlin2019bert, radford2019language, touvron2023llama, guo2025deepseek} have revolutionized the field of natural language processing (NLP). The transformer architecture~\citep{vaswani2017attention}, with its self-attention mechanism, serves as the foundation for these models. Early approaches employed the encoder-decoder structure~\citep{devlin2019bert}, but recent advancements have demonstrated the effectiveness of decoder-only architectures. Models such as GPT-2~\citep{radford2019language} and GPT-3~\citep{brown2020language} utilize next-token prediction as their core training objective, where the model learns to predict the subsequent token given a sequence of preceding tokens. This autoregressive approach has proven remarkably effective, enabling models to generate coherent and contextually appropriate text across diverse domains. Following these successes, a wave of open-source models including LLaMA~\citep{touvron2023llama}, Mistral~\citep{brown2020language}, and deepseek~\citep{guo2025deepseek} have adopted similar training paradigms while introducing architectural refinements to enhance efficiency and performance. The next-token prediction objective has emerged as a surprisingly powerful learning signal, enabling these models to capture complex linguistic patterns, factual knowledge, and even rudimentary reasoning capabilities without explicit supervision for these skills.

\subsection{Image Generation.}
Generative adversarial networks (GANs)~\citep{goodfellow2014generative, brock2018large, karras2019style, kang2023scaling} are the pioneering method for visual generation in the deep learning era, focusing on learning to generate realistic images through adversarial training. 
More recently, Diffusion models~\citep{song2019generative,ho2020denoising,dhariwal2021diffusion,lu2022dpm,rombach2022high} introduce a novel approach, treating visual generation as a reverse diffusion process, where images are gradually denoised from Gaussian noise through a series of steps. These models have demonstrated exceptional capabilities in image synthesis, particularly for high-resolution and photorealistic outputs.

\textbf{Autoregressive Image Generation.} 
In autoregressive image generation, 2D visual data is transformed into 1D sequences of either pixels or tokens, with each element generated sequentially following a predetermined ordering.
Pioneer studies \citep{van2016pixel, van2016conditional, parmar2018image, chen2020generative} have successfully validated the efficacy of autoregressive frameworks for RGB pixel generation, yielding results that rival those of GANs \citep{goodfellow2014generative,brock2018large,karras2019style,kang2023scaling}.
The introduction of VQVAE \citep{vqvae} and its hierarchical extension VQVAE-2 \citep{vqvae2, razavi2019generating} established methods for representing images as discrete token sequences in latent space. Building on this foundation, VQ-GAN \citep{esser2021taming} incorporated adversarial training to enhance perceptual quality, while RQ-Transformer \citep{lee2022autoregressive} further refined autoregressive learning within these discrete representation spaces.
Furthermore, drawing inspiration from BERT \citep{devlin2018bert}, models incorporating masked prediction techniques \citep{chang2022maskgit, yu2023magvit, yu2023magvit, yu2023language} have emerged, enabling parallel prediction of multiple tokens per iteration, thereby substantially reducing computational demands during deployment.

Recently, by expanding vocabulary size \citep{llamagen, yu2023language} and improving training strategy~\citep{pang2024randar, yu2024randomized}, autoregressive image generation methods \citep{llamagen, var, li2024autoregressive} have shown promising performance, exceeding methods based on diffusion \citep{song2019generative,ho2020denoising,dhariwal2021diffusion,lu2022dpm,rombach2022high} on several benchmarks.
Moreover, similar to our findings, concurrent work~\citep{hu2025improving} also discovers that similar codewords have similar effects on the resulting images. Based on this observation, they propose an additional training loss to guide the prediction toward the correct token class cluster. Different from their approach, we propose to predict tokens from coarse to fine, which not only brings more performance improvement but also helps to improve the sampling speed.

\section{Conclustion}

In this work, we present a novel coarse-to-fine token prediction framework for autoregressive image generation, which helps to alleviate the vocabulary redundancy problem in large VQ-VAE codebooks.
Extensive experiments demonstrated that our method substantially outperforms baseline approaches, improving Inception Scores by an average of 59 points.
Moreover, despite introducing an additional prediction step, our method accelerates the sampling process, making it both more effective and more efficient than conventional approaches.

{
    \small
    \bibliographystyle{ieeenat_fullname}
    \bibliography{main}
}

\end{document}